\documentclass[11pt,a4paper]{article}
\usepackage[hyperref]{acl2019}
\usepackage{times}
\usepackage{latexsym}
\usepackage{booktabs}
\usepackage{amsmath}
\usepackage{makecell}
\usepackage{subcaption}
\usepackage{graphicx}
\usepackage{amsmath,amssymb}
\usepackage{hyperref}

\aclfinalcopy %

\title{Learning Household Task Knowledge from WikiHow Descriptions}

\author{Yilun Zhou \\
  MIT CSAIL \\
  \tt yilun@\\\tt mit.edu \\\And
  Julie A. Shah \\
  MIT CSAIL \\
  \tt julie\_a\_shah@\\\tt csail.mit.edu \\\And
  Steven Schockaert \\
  Cardiff University \\
  \tt SchockaertS1@\\\tt cardiff.ac.uk \\}

\date{}

\begin{document}
\maketitle

\begin{abstract}
  Commonsense procedural knowledge is important for AI agents and robots that operate in a human environment. 
  While previous attempts at constructing procedural knowledge are mostly rule- and template-based, recent advances in deep learning provide the possibility of acquiring such knowledge directly from natural language sources. 
  As a first step in this direction, we propose a model to learn embeddings for tasks, as well as the individual steps that need to be taken to solve them, based on WikiHow\footnote{\url{https://www.wikihow.com}} articles. We learn these embeddings such that they are predictive of both step relevance and step ordering. We also experiment with the use of integer programming for inferring consistent global step orderings from noisy pairwise predictions.  %

\end{abstract}

\section{Introduction}
For AI agents to serve as competent (digital or physical) assistants in everyday environments, they need an understanding of the common tasks that people perform.
In contrast to factual knowledge, which is encoded to some extent in knowledge graphs such as Freebase \cite{bollacker2008freebase}, there are currently no resources that capture such knowledge in a comprehensive way.

As a natural solution, in this paper, we consider the problem of learning procedural knowledge from text descriptions, focusing on household tasks as a case study. There are two reasons for this particular focus. First, household tasks require a rich amount of commonsense knowledge, which makes them challenging to deal with for AI agents. Second, learning such knowledge has important applications in the context of household robots and smart home technologies, among others. 

The biggest challenge associated with household tasks is the lack of explicit structured information. While specialized datasets for some aspects of household tasks are available (e.g.\ cooking recipes \cite{yagcioglu2018recipeqa}, in-home navigation commands \cite{matuszek2013learning}, human action trajectories for chores \cite{koppula2013learning}), general information only exists in natural language format as descriptions intended for human consumption. With recent advances in deep learning and text mining, it is natural to wonder whether, and to what extent, we can acquire knowledge about household tasks from existing textual sources. To start answering this question, in this paper we tap into WikiHow, one of the largest online databases of procedural knowledge. Our aim is to jointly learn two types of knowledge: (i) whether a certain step pertains to a certain task and (ii) how to order two (potentially non-sequential) steps for a given task. We evaluate our learned model both in terms of the performance achieved on these two tasks and by analyzing the resulting embeddings.

\section{Related Work}
\paragraph{Knowledge Representation}
A large number of knowledge graphs have already been constructed, capturing a wide variety of human knowledge.
These graphs, such as Freebase \cite{bollacker2008freebase} and ConceptNet \cite{liu2004conceptnet}, all share the common structure of using nodes to represent concepts and using edges to represent relations. Among many applications, \citet{williams2017understanding} showed that ConceptNet can enable better story understanding by capturing some aspects of commonsense knowledge. 

However, there is little procedural knowledge in ConceptNet. Instead, planning approaches have traditionally been used to model such knowledge. In classical planning languages, such as PDDL \cite{mcdermott1998pddl}, the environment is described with a set of predicates and actions are defined in terms of pre-conditions and post-conditions. This turns planning into a search problem. While efficient and provably optimal in small domains, it is hard to model the full spectrum of real world environments with such exact definitions. By contrast, in our work we take a complementary approach, acquiring implicit knowledge from large amounts of data.

\paragraph{Embedding learning}
Vector space embeddings are commonly used to represent the semantics of linguistic constructs such as words and sentences as vectors in a high-dimensional space. At word level, embedding models such as word2vec \cite{mikolov2013distributed} and GloVe \cite{pennington2014glove} are trained based on linguistic context, i.e.\ the representation of a word depends on the words surrounding mentions of that word in some text corpus. At sentence level, embeddings can be trained from context or learned for one or several downstream applications \cite{radford2018improving, devlin2018bert}. 

Embeddings for various other kinds of data have also been studied. For example, \citet{chung2018speech2vec} learn vector representations from audio data. Moreover, a large number of methods for embedding graph and network structures have been proposed \cite{goyal2018graph}. 

\paragraph{Script knowledge}
In our work, we learn embeddings for natural language instructions targeted to facilitate commonsense knowledge acquisition. 
One of the ways in which such embeddings can be used is to rank step instructions for a specific task. Specifically, given the name of the task and two steps, all expressed in natural language, we want our model to predict which step should be done first, using only the embeddings of the task name and steps as input. This problem falls into the general category of learning script knowledge, for which several models have already been proposed. For example, \citet{chambers2009unsupervised} proposed one of the first models to learn script knowledge based on estimated mutual information between events. \citet{modi2014inducing} learned embeddings for events such that a linear ranking function operating on embedding space can be used to infer event orders. \citet{pichotta2016learning} learned orders from parsed event representation with a LSTM model. \citet{pichotta2016using} used a sentence-level LSTM model that does not require explicit event parsing and extraction.

\paragraph{Knowledge Acquisition from WikiHow}
There have been many works on collecting knowledge from the web. With specific focus on WikiHow, \citet{chu2017distilling} used information retrieval and embedding-based clustering to distill a knowledge base of task execution. By using the OpenIE system \cite{angeli2015leveraging}, they inferred relations between tasks and steps so that the distilled knowledge base recognizes that the task in a WikiHow article $A$ is equivalent to a step $B$ in another WikiHow article $C$, and the user, when reading article $C$ can look up detailed instructions for step $B$ by reading the automatically linked article $A$. In this way, a hierarchical structure among articles can be extracted. 

\citet{park2018learning} used a neural network model to learn specific relations between the steps of each task. Specifically, three relations \tt is\_method\_of\rm , \tt is\_subtask\_of\rm, and \tt is\_alternative\_of \rm are learned using a hierarchical attention neural network that achieved superior performance than standard approaches using an information extraction pipeline.

\section{Method}

\subsection{Dataset Description}

We collected a corpus consisting of all WikiHow articles under the category of ``Home and Garden\footnote{\url{https://www.wikihow.com/Category:Home-and-Garden}}'', which we believe is most relevant for our purpose of understanding household procedural knowledge. Each WikiHow article describes a particular task, which is composed of a number of steps. Some example tasks are shown in Table \ref{titles}. Each step is represented by a \emph{gist} and an \emph{explanation}. The gist is a brief and concise summary of the step, such as ``purchase packing supplies''. The corresponding explanation gives additional contexts and details to the gist. For the previous example, the explanation is as follows:
\begin{quote}
``Furniture should generally not be placed in a truck without wrapping it in some sort of protective material. After you've completed your inventory, consider what you'll need to move each piece of furniture...''
\end{quote}
Some additional examples of step gists as shown in Table \ref{steps}.

\begin{table} %
    \centering
    \begin{tabular}{l}\toprule
Remove Staples with Your Bare Hands\\\hline
Buy a Shipping Container\\\hline
Pick Up Broken Glass Splinters\\\hline
Clean Fireplace Glass\\\hline
Clean an Espresso Machine\\\bottomrule
    \end{tabular}
    \caption{A random sample of task titles}
    \label{titles}
\end{table}

\begin{table} %
    \centering
    \begin{tabular}{l}\toprule
        Shake your clothes\\\hline
Move the bowl\\\hline
Dig a hole about 2 ft deep\\\hline
Take out the trash\\\hline
Steam clean older carpets\\\bottomrule
    \end{tabular}
    \caption{A random sample of task steps}
    \label{steps}
\end{table}

An additional particularity with WikiHow is that there are two types of article structures. About 30\% of all articles have flat structures, which means that an article has a list of individual steps. The remaining ones have 2-level hierarchical structures, which means that an article has several titled subsections, and each subsection has a list of individual steps. For example, an article on ``clean kitchen'' may include subsections on ``organize kitchen shelves'', ``clean countertop'', and ``remove oil stain on floor''. We found that subsection titles are semantically and syntactically very similar to article titles, so we simply consider each subsection as a separate task. With this preprocessing, the dataset contains 12,431 articles with a total of 162,771 individual steps.\footnote{The dataset and the model implementation can be downloaded at \url{https://github.com/YilunZhou/wikihow-embedding/}}

\subsection{Model Architecture}
A task title $t$ is a list of $l$ natural language words $(w_{1}, ..., w_{l})$. For each word, we use the pre-trained GloVe embedding \cite{pennington2014glove} to look up its vector representation. This embedding is fixed during training. Words which are not in the GloVe vocabulary are represented using a special $<$unk$>$ token, the embedding for which is learned. We use $\vec v_i$ to denote the embedding corresponding to the word $w_i$. Hence the task title $t$ is represented as $t=(\vec v_1, ..., \vec v_l)$. The representations for step gists and step explanations are analogous. 

\begin{figure} %
    \centering
    \begin{subfigure}[b]{\linewidth}
    \centering
    \includegraphics[width=0.8\linewidth]{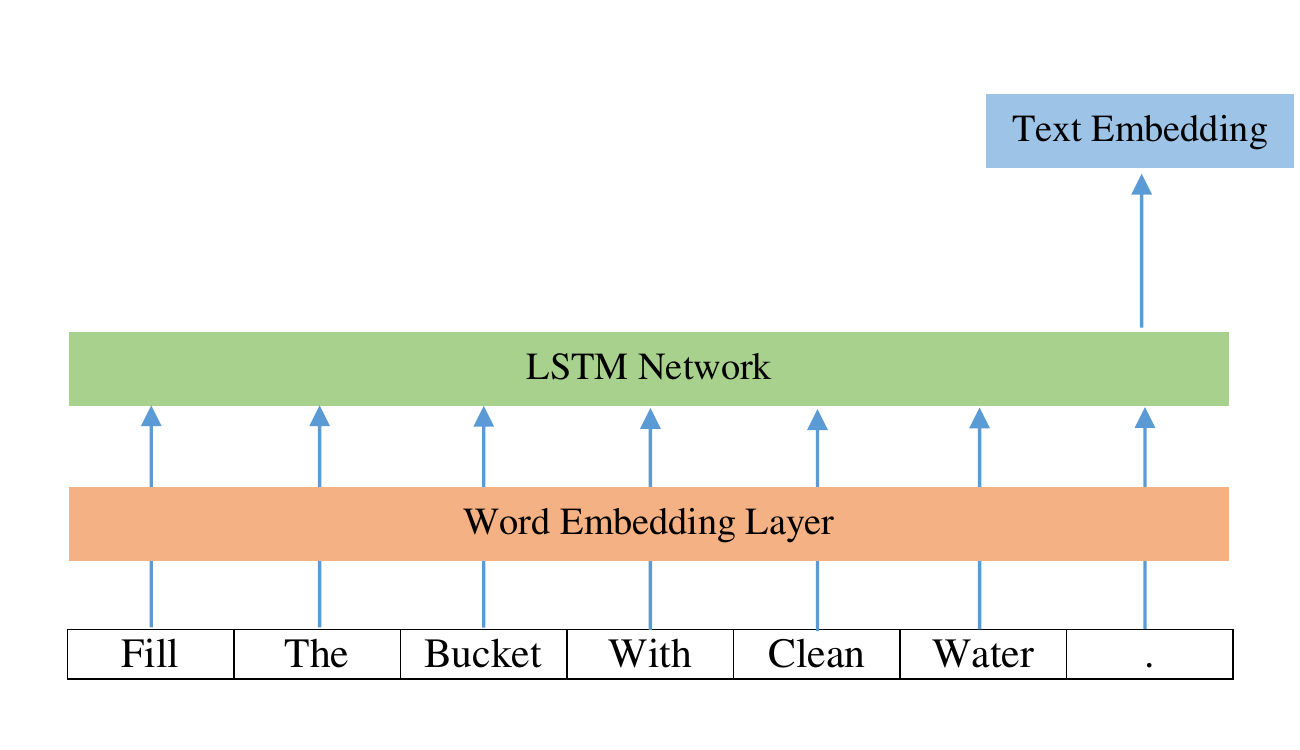}
    \caption{Embedding of task title, step gist, and step explanation}
    \end{subfigure}
    \begin{subfigure}[b]{\linewidth}
    \centering
    \includegraphics[width=0.8\linewidth]{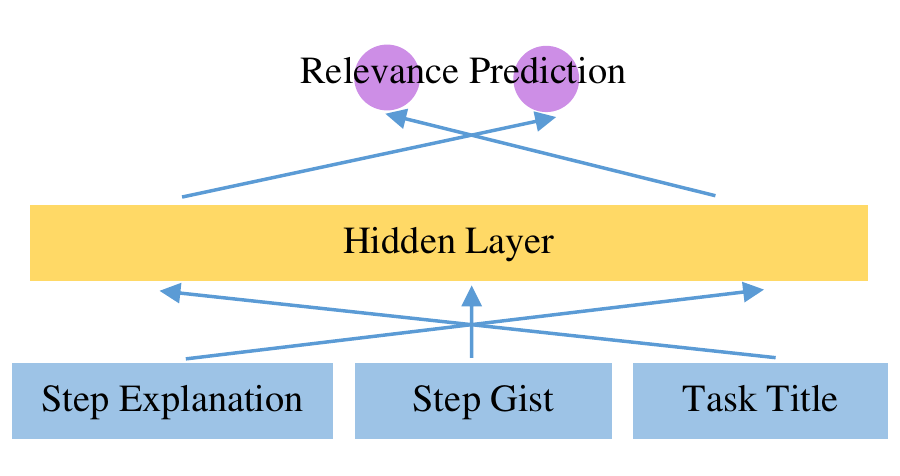}
    \caption{Step relevance prediction}
    \end{subfigure}
    \begin{subfigure}[b]{\linewidth}
    \centering
    \includegraphics[width=\linewidth]{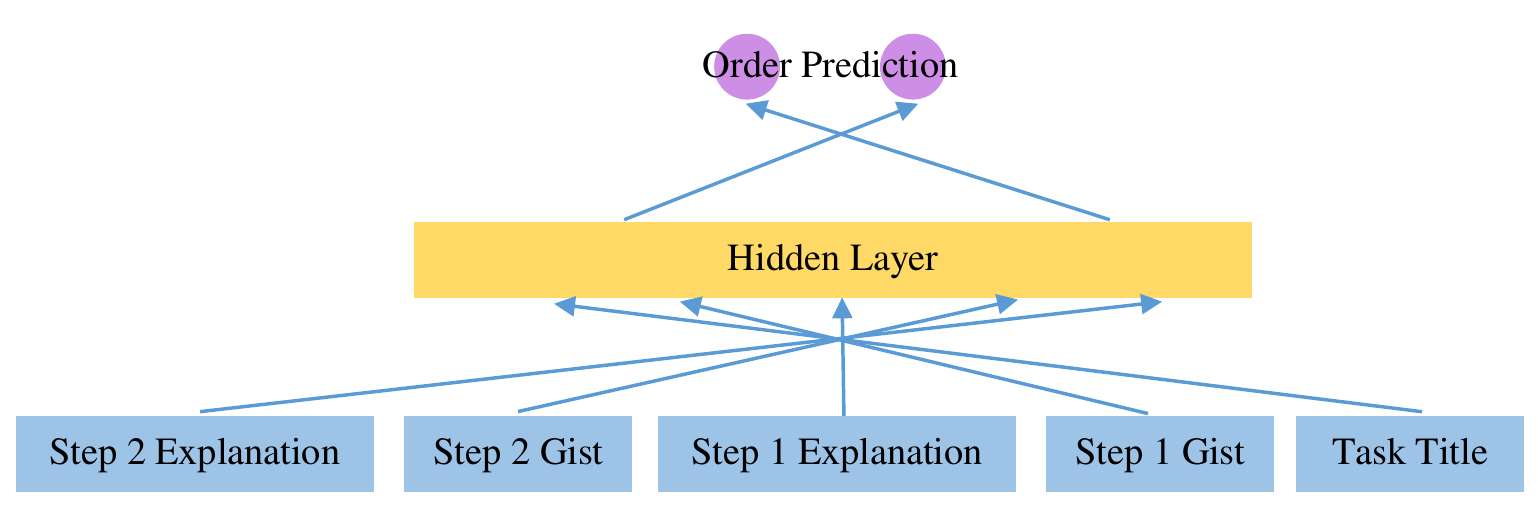}
    \caption{Step order prediction}
    \end{subfigure}
    \caption{Model architecture}
    \label{model}
\end{figure}

Figure \ref{model} shows the general workflow of our model. First, three LSTM networks are used to encode the task titles, step gists and step explanations. The input to each of these LSTMs is a list of word vectors, as explained above, and the outputs are the corresponding embeddings, which are taken as the last hidden state of the LSTM encoder. The initial hidden units and memory units of the LSTMs are initialized to 0 (and not updated during training).

\subsection{Step Relevance Prediction}
Predicting the relevance of a step to a task is a binary classification problem. We first concatenate the embeddings for the task title, step gist, and step explanation together to form the input vector. Then this vector is passed through a hidden layer and an output layer to get the probability that the step is relevant to the task. Negative log-likelihood (NLL) loss is used during training. 

\subsection{Step Ranking Prediction}
Given the task name and two steps from the WikiHow description of that task, we use a similar fully connected network to predict whether a step should happen before another step from the concatenation of embeddings of the task and the two steps, also with NLL loss. %

\subsection{Joint Prediction of Step Ordering}
Given an unordered set of steps, for a given task, the aforementioned neural network can be applied to make a prediction for pairwise step orderings. However, since these predictions are made independently, they may be conflicting with each other
(e.g. A is predicted before B, B is predicted before C, and C is predicted before A). 
Furthermore, we recognize that sometimes two steps can be done in parallel, and a penalty should not be incurred for incorrectly predicting the ordering in which these two steps happen to be ordered in WikiHow. Thus, to be fully flexible with the possibility of ``ambiguous'' ordering, we employ an integer programming (IP) formulation. 

For each pair of steps $(i,j)$, with $i\neq j$, we introduce two binary variables $x_{ij}$ and $x_{ji}$. The meaning of $x_{ij}=1$ is that step $i$ has to occur (strictly) before step $j$, while $x_{ij}=0$ means that either step $i$ has to occur (strictly) after step $j$ (if $x_{ji}=1$), or that the ordering does not matter (if $x_{ji}=0$). Then we set up the following IP problem: 
\begin{align*}
 \mbox{maximize } &\sum_{i, j} w_{ij}x_{ij}, \\
 \mbox{subject to } & x_{ij}\in\{0, 1\} \;\;\;\;\;\forall i, j, \\
 &x_{ij} + x_{ji} \leq 1 \;\;\;\;\;\forall i, j, \\
 &x_{ij} + x_{jk} - x_{ik} \leq 1 \;\;\;\;\;\forall i, j, k, \\
 &\sum_{i,j} x_{ij} \geq D. 
\end{align*}
In the objective function, we choose $w_{ij}=\log\allowbreak \Pr(i \mbox{ before } j)$, where this log probability is predicted by the neural network. The first constraint enforces the binary nature of $x_{ij}$. The second constraint requires that $x_{ij}$ and $x_{ji}$ cannot be both 1, as that would mean that step $i$ is both strictly before and after step $j$, which is not possible. The third constraint enforces transitivity: if step $i$ is strictly before step $j$, and step $j$ is also strictly before step $k$, then we must have that step $i$ is strictly before step $k$. At this stage, it is easy to see that since $w_{ij}\leq 0$ due to $w_{ij}$ being a log probability, the optimal solution is achieved by choosing $x_{ij}=0$ for each $i$ and $j$. Indeed, no penalty is incurred if all pairwise relations are predicted to be ambiguous. For this reason, we have the final constraint which imposes that at least $D$ pairs should be ordered. Note that for a task with $T$ steps, $D$ can be at most $T(T-1)/2$ (i.e.\ half of all total pairs). Otherwise the second constraint would be unsatisfiable. 

\subsection{Learning and Inference}
We used PyTorch \cite{paszke2017automatic} to implement the feed-forward and back-propagation of training, with Adam \cite{kingma2014adam} as the optimizer. To solve the integer programming problem, we used CVXPY \cite{diamond2016cvxpy}. 

\section{Experiments}
\subsection{Data Preparation}
We used a 80\%/10\%/10\% split of training, validation, and test data. All reported statistics are from the test set, which is held out during training. Table \ref{main-result} summarizes the results. 

For step relevance prediction tasks, we collected each positive example by sampling a task title and a random step associated with the task. For negative examples, we sampled task titles and steps independently and made sure that the step does not belong to the task. The number of positive and negative examples are balanced. 

For step ordering, for each example we sample a task and two steps. Then we randomly denote one of them as step 1 and the other as step 2, and set up the label accordingly. 
\subsection{Training Details}
We used 500-dimensional embeddings throughout, but we found that the learning performance is not sensitive to the embedding dimension, as long as it is over 100. We zero-initialized the hidden and memory cells of the LSTM encoders. The learning rate for the Adam optimizer was set to 0.001. 

\subsection{Learning Performance}

\begin{table} %
    \centering
    \begin{tabular}{c|c|c}\toprule
        & Step relevance & Step ranking \\\midrule
        \makecell{LSTM step\\explanation} & 0.911 & 0.752 \\\midrule
        \makecell{bag step\\explanation} & 0.902 & 0.664 \\\midrule
        \makecell{no step\\explanation} & 0.844 & 0.657 \\\bottomrule
    \end{tabular}
    \caption{Model performance on two prediction problems}
    \label{main-result}
\end{table}

Our model performance is summarized in the first row. In addition, we also tried directly using a bag of word representation as the representation for step explanation, while still keeping the LSTM encoder for step gist (second row). Specifically, the embedding of the step explanation is calculated as the average of all embedding vectors for words in the explanation. We also tried not using step explanation information at all, whose performance is shown in the third row.

We see that for both relevance and ranking predictions, the full model with step explanation encoded by an LSTM model performs the best. However, using a bag of words vector representation of step explanations still performs better than not using the step explanation at all, although the improvement is small in the case of step ranking prediction.

\begin{figure} %
    \centering
    \includegraphics[width=0.9\linewidth]{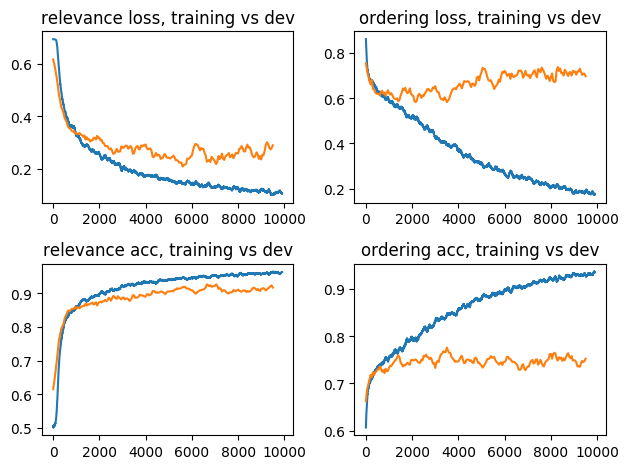}
    \caption{Training (blue) and validation (orange) loss and accuracy for two tasks}
    \label{learning_curve}
\end{figure}

The learning curve in Figure \ref{learning_curve} shows that while the model is able to get very high accuracy on the training set, the validation accuracy stabilizes after a few thousand iterations, indicating that the model is overfitting to the training set afterwards. The test performance in Table \ref{main-result} was calculated on the test set using the model iteration that achieves the highest validation accuracy. 

\subsection{Integer Programming Inference}
\label{ipresult}
\begin{figure} %
    \centering
    \includegraphics[width=0.9\linewidth]{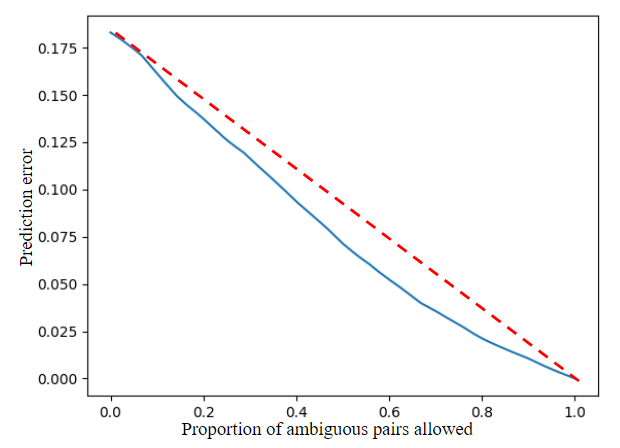}
    \caption{Rank order inference using integer programming}
    \label{ip}
\end{figure}

In this section, we study if using integer programming inference can provide better ordering performance if we allow the ordering among some pairs to be undecided (i.e.\ if we set $D$ to be strictly less than half the total number of pairs). Figure \ref{ip} presents the result. 

The horizontal axis shows the proportion of ambiguous pairs allowed, and the vertical axis shows the proportion of ordering errors. For comparison, we would achieve the red dashed line if we randomly mark pairs as ambiguous, and thus not penalized. We can see that the integer programming inference method is indeed better at identifying ambiguous pairs that, when marked as such, would lead to better performance. 
However, the improvement is not very substantial, maybe because at training time, the neural network tries to satisfy ambiguous pairs in a way that is more or less arbitrarily defined by the training data, bringing the overall performance down. Thus, one specific idea for future work would be to allow the neural network to intentionally make ambiguous predictions, for which it would not be penalized. Clearly, however, some form of regularization would need to be in place to prevent the network from making the ambiguous prediction too frequently. 

\subsection{Embedding Visualization}
\begin{figure*} %
  \includegraphics[width=\textwidth]{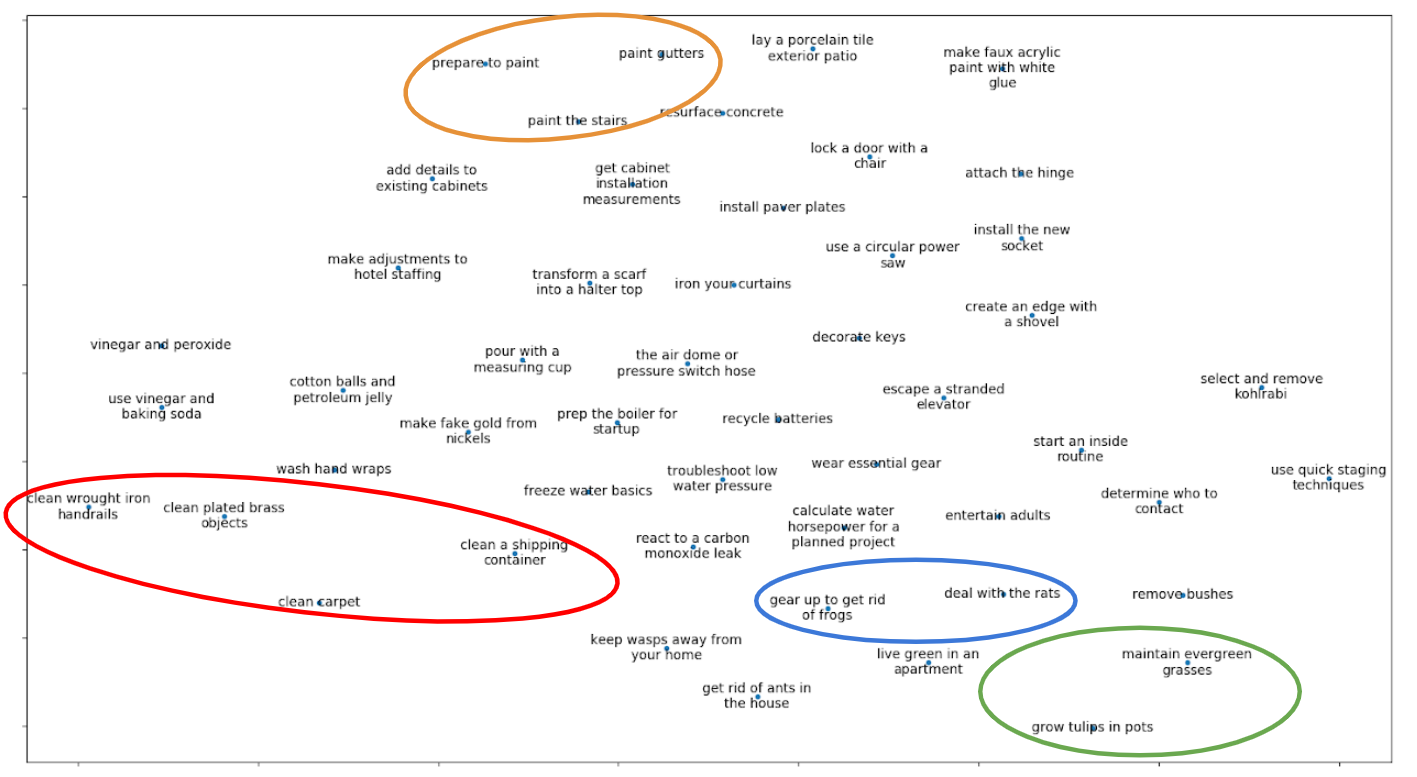}
  \caption{Embedding visualization. }
  \label{embedding}
\end{figure*}

Figure \ref{embedding} visualizes the embedding of 50 randomly selected tasks, using t-SNE \cite{maaten2008visualizing} to reduce the dimension to 2. We can see that several clusters of semantically related tasks can be identified, {which are indicated by ellipses}. 

\section{Conclusion and Future Work}

In this paper, we collected a dataset of natural language instructions for a diverse set of household tasks. We learned a joint embedding of task title and step text for two problems, predicting if a step belongs to a task title and ordering two steps given the task title. We showed that the step relevance can be predicted with a high accuracy if we include the step explanation as part of our model. However, step ordering turned out to be a more challenging problem. We believe that this is at least partly due to noise in the dataset, especially the fact that the ordering of some steps tends to be exchangeable. We also noticed some issues that are specific to WikiHow. For example, the steps which are mentioned for some tasks are more like tips and suggestions. 

In terms of future work, one direction is to ground the learned knowledge into some physical systems, such as an in-home robotic platform. Another direction is to try to recognize steps that do not have well-defined orders (Section \ref{ipresult}), although this is likely to require some additional supervision signal. 

\section{Acknowledgments}
Steven Schockaert has been supported by ERC Starting Grant 637277.

\bibliography{acl2019}

\begin{thebibliography}{24}
\expandafter\ifx\csname natexlab\endcsname\relax\def\natexlab#1{#1}\fi

\bibitem[{Angeli et~al.(2015)Angeli, Premkumar, and
  Manning}]{angeli2015leveraging}
Gabor Angeli, Melvin Jose~Johnson Premkumar, and Christopher~D Manning. 2015.
\newblock Leveraging linguistic structure for open domain information
  extraction.
\newblock In \emph{Proceedings of the 53rd Annual Meeting of the Association
  for Computational Linguistics and the 7th International Joint Conference on
  Natural Language Processing (Volume 1: Long Papers)}, volume~1, pages
  344--354.

\bibitem[{Bollacker et~al.(2008)Bollacker, Evans, Paritosh, Sturge, and
  Taylor}]{bollacker2008freebase}
Kurt Bollacker, Colin Evans, Praveen Paritosh, Tim Sturge, and Jamie Taylor.
  2008.
\newblock Freebase: a collaboratively created graph database for structuring
  human knowledge.
\newblock In \emph{Proceedings of the 2008 ACM SIGMOD international conference
  on Management of data}, pages 1247--1250. AcM.

\bibitem[{Chambers and Jurafsky(2009)}]{chambers2009unsupervised}
Nathanael Chambers and Dan Jurafsky. 2009.
\newblock Unsupervised learning of narrative schemas and their participants.
\newblock In \emph{Proceedings of the Joint Conference of the 47th Annual
  Meeting of the ACL and the 4th International Joint Conference on Natural
  Language Processing of the AFNLP: Volume 2-Volume 2}, pages 602--610.
  Association for Computational Linguistics.

\bibitem[{Chu et~al.(2017)Chu, Tandon, and Weikum}]{chu2017distilling}
Cuong~Xuan Chu, Niket Tandon, and Gerhard Weikum. 2017.
\newblock Distilling task knowledge from how-to communities.
\newblock In \emph{Proceedings of the 26th International Conference on World
  Wide Web}, pages 805--814. International World Wide Web Conferences Steering
  Committee.

\bibitem[{Chung and Glass(2018)}]{chung2018speech2vec}
Yu-An Chung and James Glass. 2018.
\newblock Speech2vec: A sequence-to-sequence framework for learning word
  embeddings from speech.
\newblock \emph{arXiv preprint arXiv:1803.08976}.

\bibitem[{Devlin et~al.(2018)Devlin, Chang, Lee, and
  Toutanova}]{devlin2018bert}
Jacob Devlin, Ming-Wei Chang, Kenton Lee, and Kristina Toutanova. 2018.
\newblock Bert: Pre-training of deep bidirectional transformers for language
  understanding.
\newblock \emph{arXiv preprint arXiv:1810.04805}.

\bibitem[{Diamond and Boyd(2016)}]{diamond2016cvxpy}
Steven Diamond and Stephen Boyd. 2016.
\newblock Cvxpy: A python-embedded modeling language for convex optimization.
\newblock \emph{The Journal of Machine Learning Research}, 17(1):2909--2913.

\bibitem[{Goyal and Ferrara(2018)}]{goyal2018graph}
Palash Goyal and Emilio Ferrara. 2018.
\newblock Graph embedding techniques, applications, and performance: A survey.
\newblock \emph{Knowledge-Based Systems}, 151:78--94.

\bibitem[{Kingma and Ba(2014)}]{kingma2014adam}
Diederik~P Kingma and Jimmy Ba. 2014.
\newblock Adam: A method for stochastic optimization.
\newblock \emph{arXiv preprint arXiv:1412.6980}.

\bibitem[{Koppula and Saxena(2013)}]{koppula2013learning}
Hema Koppula and Ashutosh Saxena. 2013.
\newblock Learning spatio-temporal structure from rgb-d videos for human
  activity detection and anticipation.
\newblock In \emph{International conference on machine learning}, pages
  792--800.

\bibitem[{Liu and Singh(2004)}]{liu2004conceptnet}
Hugo Liu and Push Singh. 2004.
\newblock Conceptnet—a practical commonsense reasoning tool-kit.
\newblock \emph{BT technology journal}, 22(4):211--226.

\bibitem[{Maaten and Hinton(2008)}]{maaten2008visualizing}
Laurens van~der Maaten and Geoffrey Hinton. 2008.
\newblock Visualizing data using t-sne.
\newblock \emph{Journal of machine learning research}, 9(Nov):2579--2605.

\bibitem[{Matuszek et~al.(2013)Matuszek, Herbst, Zettlemoyer, and
  Fox}]{matuszek2013learning}
Cynthia Matuszek, Evan Herbst, Luke Zettlemoyer, and Dieter Fox. 2013.
\newblock Learning to parse natural language commands to a robot control
  system.
\newblock In \emph{Experimental Robotics}, pages 403--415. Springer.

\bibitem[{McDermott et~al.(1998)McDermott, Ghallab, Howe, Knoblock, Ram,
  Veloso, Weld, and Wilkins}]{mcdermott1998pddl}
Drew McDermott, Malik Ghallab, Adele Howe, Craig Knoblock, Ashwin Ram, Manuela
  Veloso, Daniel Weld, and David Wilkins. 1998.
\newblock Pddl-the planning domain definition language.

\bibitem[{Mikolov et~al.(2013)Mikolov, Sutskever, Chen, Corrado, and
  Dean}]{mikolov2013distributed}
Tomas Mikolov, Ilya Sutskever, Kai Chen, Greg~S Corrado, and Jeff Dean. 2013.
\newblock Distributed representations of words and phrases and their
  compositionality.
\newblock In \emph{Advances in Neural Information Processing Systems (NIPS)},
  pages 3111--3119.

\bibitem[{Modi and Titov(2014)}]{modi2014inducing}
Ashutosh Modi and Ivan Titov. 2014.
\newblock Inducing neural models of script knowledge.
\newblock In \emph{Proceedings of the Eighteenth Conference on Computational
  Natural Language Learning}, pages 49--57.

\bibitem[{Park and Motahari~Nezhad(2018)}]{park2018learning}
Hogun Park and Hamid~Reza Motahari~Nezhad. 2018.
\newblock Learning procedures from text: Codifying how-to procedures in deep
  neural networks.
\newblock In \emph{Companion of the The Web Conference 2018 on The Web
  Conference 2018}, pages 351--358. International World Wide Web Conferences
  Steering Committee.

\bibitem[{Paszke et~al.(2017)Paszke, Gross, Chintala, Chanan, Yang, DeVito,
  Lin, Desmaison, Antiga, and Lerer}]{paszke2017automatic}
Adam Paszke, Sam Gross, Soumith Chintala, Gregory Chanan, Edward Yang, Zachary
  DeVito, Zeming Lin, Alban Desmaison, Luca Antiga, and Adam Lerer. 2017.
\newblock Automatic differentiation in pytorch.

\bibitem[{Pennington et~al.(2014)Pennington, Socher, and
  Manning}]{pennington2014glove}
Jeffrey Pennington, Richard Socher, and Christopher Manning. 2014.
\newblock {GloVe}: Global vectors for word representation.
\newblock In \emph{Proceedings of the Conference on Empirical Methods in
  Natural Language Processing (EMNLP)}, pages 1532--1543.

\bibitem[{Pichotta and Mooney(2016{\natexlab{a}})}]{pichotta2016learning}
Karl Pichotta and Raymond~J Mooney. 2016{\natexlab{a}}.
\newblock Learning statistical scripts with lstm recurrent neural networks.
\newblock In \emph{Thirtieth AAAI Conference on Artificial Intelligence}.

\bibitem[{Pichotta and Mooney(2016{\natexlab{b}})}]{pichotta2016using}
Karl Pichotta and Raymond~J Mooney. 2016{\natexlab{b}}.
\newblock Using sentence-level lstm language models for script inference.
\newblock \emph{arXiv preprint arXiv:1604.02993}.

\bibitem[{Radford et~al.(2018)Radford, Narasimhan, Salimans, and
  Sutskever}]{radford2018improving}
Alec Radford, Karthik Narasimhan, Tim Salimans, and Ilya Sutskever. 2018.
\newblock Improving language understanding by generative pre-training.

\bibitem[{Williams et~al.(2017)Williams, Lieberman, and
  Winston}]{williams2017understanding}
Bryan Williams, Henry Lieberman, and Patrick~H Winston. 2017.
\newblock Understanding stories with large-scale common sense.
\newblock In \emph{COMMONSENSE}.

\bibitem[{Yagcioglu et~al.(2018)Yagcioglu, Erdem, Erdem, and
  Ikizler-Cinbis}]{yagcioglu2018recipeqa}
Semih Yagcioglu, Aykut Erdem, Erkut Erdem, and Nazli Ikizler-Cinbis. 2018.
\newblock Recipeqa: A challenge dataset for multimodal comprehension of cooking
  recipes.
\newblock \emph{arXiv preprint arXiv:1809.00812}.

\end{thebibliography}
\bibliographystyle{acl_natbib}

\end{document}